\pdfoutput=1
\documentclass[11pt]{article}

\usepackage[preprint]{coling}

\usepackage{times}
\usepackage{latexsym}

\usepackage[T1]{fontenc}
\usepackage[utf8]{inputenc}

\usepackage{microtype}

\usepackage{inconsolata}

\usepackage{graphicx}

\usepackage{times}
\usepackage{latexsym}
\usepackage{url}
\usepackage{booktabs}

\usepackage{graphicx}
\usepackage{xcolor}
\usepackage{graphicx}
\usepackage{caption}
\usepackage{subcaption}
\usepackage{mathtools, nccmath}
\usepackage{multirow}
\usepackage{caption}
\usepackage{subcaption}
\usepackage{enumitem,amssymb}
\usepackage{pifont}

\usepackage[T1]{fontenc}
\usepackage[utf8]{inputenc}

\usepackage{microtype}
\usepackage{amsmath}
\usepackage{inconsolata}

\title{Exploring Large Language Models for Product Attribute Value Identification}

\author{
    Kassem Sabeh\textsuperscript{1}, Mouna Kacimi\textsuperscript{2}, Johann Gamper\textsuperscript{1}, 
    Robert Litschko\textsuperscript{3}, Barbara Plank\textsuperscript{3} \\
    \textsuperscript{1}Free University of Bozen-Bolzano, Italy, \{ksabeh, jgamper\}@unibz.it \\
    \textsuperscript{2}Wonder Technology Srl, Italy, mouna@wonderflow.ai \\
    \textsuperscript{3}LMU Munich, Germany, \{robert.litschko, b.plank\}@lmu.de
}

\begin{document}
\maketitle
\begin{abstract}
Product attribute value identification (PAVI) involves automatically identifying attributes and their values from product information, enabling features like product search, recommendation, and comparison. Existing methods primarily rely on fine-tuning pre-trained language models, such as BART and T5, which require extensive task-specific training data and struggle to generalize to new attributes. This paper explores large language models (LLMs), such as LLaMA and Mistral, as data-efficient and robust alternatives for PAVI. We propose various strategies: comparing one-step and two-step prompt-based approaches in zero-shot settings and  utilizing parametric and non-parametric knowledge through in-context learning examples. We also introduce a dense demonstration retriever based on a pre-trained T5 model and perform instruction fine-tuning to explicitly train LLMs on task-specific instructions. Extensive experiments on two product benchmarks show that our two-step approach significantly improves performance in zero-shot settings, and instruction fine-tuning further boosts performance when using training data, demonstrating the practical benefits of using LLMs for PAVI.
\end{abstract}

\section{Introduction}

Product attributes are essential elements of e-commerce platforms that significantly enhance various applications, benefiting both customers and the platform itself. They provide detailed information about product features, enabling customers to compare products and make informed purchasing decisions \citep{ren2018information}. For e-commerce platforms, product attributes are crucial for powering applications such as product search \cite{nguyen2020learning,chen2023generate}, recommendation systems \cite{yu2021leveraging,truong2022ampsum}, and product-related question answering \cite{huang2022autoregressive,rozen2021answering}. This dual importance for both customers and e-commerce platforms has established product attribute value identification (PAVI) as a core task in the e-commerce industry \citep{zheng2018opentag,shinzato2023unified}.

\begin{figure}
    \centering \includegraphics[width=0.9\columnwidth]{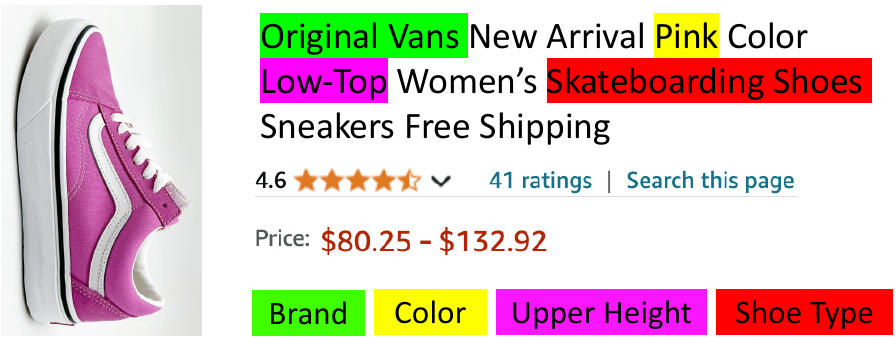}
    \caption{An example of a product listing with tagged attribute-value pairs in the title.}
    \label{fig:example}
\end{figure}

PAVI refers to the task of identifying both the attributes and their corresponding values from an input context, such as a product title or description\footnote{\citet{roy2024exploring} refers to the PAVI task as joint attribute and value extraction.}. For example, given the product title in Figure \ref{fig:example}: "Original Vans New Arrival Pink Color
Low-Top Women’s Skateboarding Shoes
Sneakers Free Shipping", a model should identify the attributes Brand, Color, Upper Height and Shoe Type, with the corresponding values Original Vans, Pink, Low-Top, and Skateboarding Shoes.

Most existing work focuses on Product Attribute Value Extraction (PAVE), which simplifies the task as a sequence tagging \cite{putthividhya2011bootstrapped,zheng2018opentag,xu2019scaling} or question-answering \cite{wang2020learning,yang2022mave,sabeh2024qpave} problem, where attribute names are provided as part of the input, and the goal is to extract the corresponding values. However, these methods need to be applied repeatedly for thousands of attributes, making them impractical without a comprehensive and accurate attribute taxonomy, which is often unavailable or imperfect in real-world settings \cite{shinzato2023unified, mao2020octet}. With e-commerce platforms featuring over ten thousand attributes \cite{xu2019scaling}, this approach becomes computationally expensive and difficult to scale. In contrast, PAVI is a more realistic and complex task, requiring the generation of attribute names directly from the input context, making it more challenging and representative of real-world scenarios \cite{shinzato2023unified}.

Existing work on PAVI is limited and primarily relies on fine-tuning pre-trained language models (PLMs) like T5 \cite{raffel2020exploring} and BART \cite{lewis2019bart}. Recent approaches, like those by \citet{roy2024exploring} and \citet{shinzato2023unified}, fine-tune generative models to jointly extract attributes and values from product text. These approaches, however, require large amounts of task-specific training data and struggle in zero-shot settings where no such data exists. In real-world scenarios, new products and attributes constantly emerge, requiring models to adapt quickly to unseen attributes without costly retraining. Existing methods that rely on retraining struggle in these situations, as they demand continuous updates and often fail to generalize well to new, unseen attributes \cite{yang2023mixpave}. Large auto-regressive language models (LLMs) \cite{dubey2024LLaMA,jiang2023mistral,groeneveld2024OLMo} offer a promising alternative due to their ability to generalize across domains without task-specific data, excelling in zero-shot settings with minimal fine-tuning \cite{agrawal2022large,mann2020language,wadhwa2023revisiting}. While LLMs have been explored for PAVE \cite{brinkmann2023product}, they have yet to be applied to the more complex task of PAVI.

This paper explores the potential of large language models (LLMs) for PAVI. We focus on open-source LLMs, like LLaMA-3 \cite{dubey2024LLaMA}, Mistral \cite{jiang2023mistral}, and OLMo \cite{groeneveld2024OLMo}. Unlike traditional PLMs that require extensive fine-tuning, LLMs have the advantage of performing well in zero-shot settings, where no training data is available. However, given the complexity of the PAVI task, relying solely on zero-shot capabilities may not suffice \cite{roy2024exploring}. The primary bottleneck for PAVI is attribute recall, as improving PAVI is closely tied to enhancing the model's ability to identify attributes \cite{sabeh2024empirical}. To address this, we incorporate parametric and non-parametric in-context examples to guide the models. Parametric knowledge refers to examples generated internally by the model itself, utilizing its pre-trained capabilities. Non-parametric knowledge, on the other hand, involves retrieving external data such as unlabelled product titles or labeled examples from a training set. Additionally, we explore instruction fine-tuning to explicitly train the LLMs on task-specific instructions, aiming to enhance their performance.

We propose and evaluate two methods for performing PAVI: a one-step approach and a two-step approach. In the one-step approach, the model is prompted to directly identify attribute-value pairs. In the two-step approach, the model first identifies attributes and then extracts the corresponding values. We compare different strategies for utilizing in-context examples, including (i) using product titles without labels, (ii) providing labeled examples, (iii) employing various methods to select in-context demonstrations, including a fine-tuned dense retriever, and (iv) instruction fine-tuning to explicitly train the models on the PAVI task. We evaluate the performance of these strategies on two real world product datasets: AE-110k \cite{xu2019scaling} and OA-Mine \cite{zhang2022oa}. The main contributions of this work are summarized as follows:

\begin{itemize} [left=0pt, itemsep=0pt]

    \item  We are the first to explore LLMs for PAVI, a more complex task than PAVE as it requires generating attribute names from context. We propose one-step and two-step methods, showing the two-step approach significantly improves zero-shot performance.

    % \item Our experiments demonstrate that self-generated examples improve performance in zero-shot settings.

    \item  We explore different approaches to incorporate in-context examples, using parametric and non-parametric knowledge, and show that our proposed dense retriever provides the best results. We also investigate domain transfer for dense in-context retrievers.

    \item We conduct instruction fine-tuning experiments, showing how explicit task-specific training further enhances LLM performance for PAVI.

\end{itemize}

\section{Related Work}

\paragraph{Product Attribute Value Extraction.}
Early approaches for PAVE included rule-based \cite{vandic2012faceted,gopalakrishnan2012matching} and named entity recognition (NER)-based techniques \cite{brooke2016bootstrapped,chen2019exact}, which suffer from limited coverage and closed-world assumptions. With the advent of neural networks, sequence tagging models were introduced to address these limitations \cite{huang2015bidirectional,yan2021adatag,xu2019scaling}. OpenTag \cite{zheng2018opentag}, for example, used a BiLSTM-CRF model with active learning to enhance attribute-value tagging. Recent advancements involve transformer-based architectures \cite{yan2021adatag,shinzato2022simple,yang2024eave}. For instance, methods like AVEQA \cite{wang2020learning} and MAVEQA \cite{yang2022mave} formulate the problem as a question-answering task, using BERT \cite{devlin-etal-2019-bert} to encode the target attributes, categories, and titles. Zero-shot and few-shot learning methods, such as OA-Mine \cite{zhang2022oa} and MixPAVE \cite{yang2023mixpave}, have also been explored to address the challenge of extracting attribute values with minimal annotated data. Additionally, multi-modal approaches \cite{zhu2020multimodal,lin2021pam,wang2022smartave,gong2023knowledge} further enhance the performance by integrating visual features. More recently, generative approaches leveraging LLMs have been proposed \cite{brinkmann2023product,baumann2024using,fang2024llm}. These models decode attribute values directly, offering a promising direction for future research in PAVE.

\paragraph{Product Attribute Value Identification.}
PAVI is much less studied in the literature compared to PAVE because it is a more complex task that requires the attributes to be generated and not assumed as part of the input. Early works formalized PAVI as a multi-label classification problem. \citet{chen2023generate} proposed a multi-label classification model with attribute-value pairs as target labels. This approach struggled with skewed label distribution, which they mitigated using a label masking method to reduce negative labels via an attribute taxonomy. \citet{fuchs2022product} decomposed the target label into two atomic labels, attribute and value, performing hierarchical classification to mitigate the extreme multi-class classification problem. However, these classification methods struggle with unseen values.
Recent studies have  explored generative models for PAVI. \citet{roy2024exploring} proposed a generative framework for joint attribute and value extraction, showing it outperforms question-answering methods on the AE-110k dataset \cite{xu2019scaling}. Similarly, \citet{shinzato2023unified} fine-tuned a pre-trained T5 model \cite{raffel2020exploring} to decode target attribute-value pairs from the input product text of the MAVE dataset \cite{yang2022mave}, showing that generative methods outperform traditional extraction and classification-based approaches \cite{chen2023generate}. \citet{sabeh2024empirical} further formalized PAVI as an attribute-value generation task and compared different strategies, highlighting the potential of generative models in handling this complex task.
Despite these advancements, transformer-based models require significant  computational resources and training data, limiting their use in real-world scenarios with thousands of attributes.

\paragraph{LLMs for Feature Extraction.}
Large Language Models outperform PLMs in zero-shot tasks and are more robust to unseen examples due to their extensive pre-training on vast text corpora and their large model sizes \cite{brown2020language}. LLMs have been effectively applied to various information extraction tasks across different domains \cite{xu2023large}. For example, \citet{wang2023code4struct} and \citet{parekh2023geneva} utilized OpenAI’s LLMs to extract structured event data from unstructured text sources. \citet{agrawal2022large} used InstructGPT \cite{ouyang2022training} with zero-shot and few-shot prompts to extract information from clinical notes. \citet{goel2023LLMs} applied LLMs to extract patient information from medical text. Similarly, \citet{chen2023knowledge} used LLMs to identify product relationships, and \citet{maragheh2023llm} extracted keywords for products inferred from their textual data. Furthermore, \citet{shyr2023identifying} evaluated LLMs for extracting rare disease phenotypes from text. Additionally, LLMs have been used to rank items by extracting features from prompts and re-ranking PLM-extracted information \cite{hou2024large}. This research explores the potential of LLMs for PAVI, building on the demonstrated capabilities of these models in other domains. To the best of our knowledge,
this is the first study to investigate LLMs for product attribute value identification.

\section{Methodology}

In this section, we present our methodology for PAVI, detailing the approaches used to model the task and leverage LLMs effectively.

\subsection{PAVI Task Modelling}

Product Attribute Value Identification can be naturally decomposed into two sub-tasks: attribute identification and value extraction. Based on this decomposition, PAVI can be approached either by simultaneously generating both attributes and values in a single step or by separating these tasks into distinct stages.

\paragraph{One-Step Approach.}
In the one-step approach, the model is prompted to directly extract attribute-value pairs from the input text in a single step. This method treats the identification of attributes and their corresponding values as a combined task, allowing the model to use its contextual understanding to generate structured outputs efficiently. The prompt instructs the model to recognize and output both the attribute and value simultaneously, making full use of the LLM’s generative capabilities. An example of this prompt can be seen in Figure \ref{fig:one-step-zs}.

\paragraph{Two-Step Approach. }
The two-step approach decomposes the PAVI task into two sequential stages for a more structured extraction process. In the first stage, the model is prompted to identify attributes from the input text. In the second stage, these identified attributes are used to prompt the model again, this time to extract the corresponding values. This staged prompting allows the model to refine its understanding of the task incrementally, improving the accuracy of both attribute identification and value extraction\footnote{We also experimented with reversing the sequence, extracting values first followed by attribute identification. However, preliminary experiments indicated that identifying attributes first yields better performance.}. An example template of this prompt can be found in Figure \ref{fig:two-step-zs}.

\subsection{Leveraging LLMs for PAVI}

This section explores how LLMs are utilized for PAVI in zero-shot settings, in-context learning, and instruction fine-tuning.

\begin{figure}
    \centering \includegraphics[width=1\columnwidth]{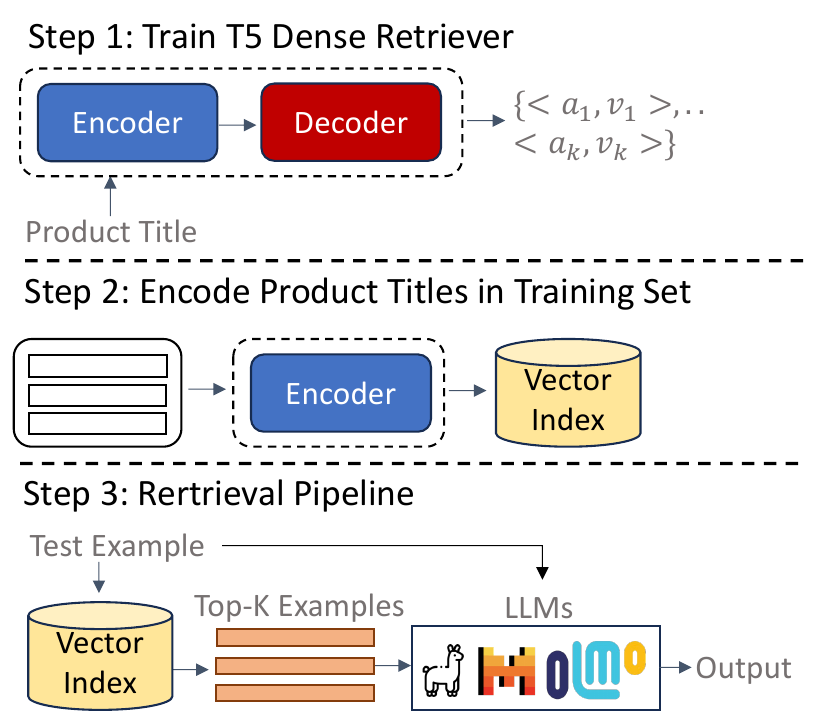}
    \caption{Proposed method for using the fine-tuned T5 model as a dense retriever.}
    \label{fig:method}
\end{figure}

\subsubsection{Zero-Shot Settings}
In zero-shot settings, the model performs PAVI without any task-specific training data, relying solely on its pre-trained knowledge. This scenario evaluates the model's ability to generalize and identify attribute-value pairs based on its extensive language understanding acquired during large-scale pre-training. To assess zero-shot performance, we utilize the one-step and two-step approaches described earlier. By comparing these approaches in zero-shot settings, we aim to determine which method more effectively leverages the model’s pre-trained knowledge to accurately identify and extract attribute-value pairs without relying on task-specific training data. This evaluation sheds light on the inherent strengths and potential limitations of each approach, offering guidance on the most effective use of LLMs for PAVI in real-world scenarios where no training data is available.

\subsubsection{In-context Learning}

In this section, we explore how in-context learning can enhance PAVI performance by using both parametric and non-parametric knowledge sources.

\paragraph{Parametric Knowledge.}
To enhance the model's performance in zero-shot settings, we leverage the model’s internal parametric knowledge by generating examples using the model itself. Specifically, given a test input, the model is prompted to generate pseudo inputs in the form of example product titles, rather than labelled examples. These generated titles act as additional context that guides the model \cite{chen2023self}. 
In the one-step approach, the generated product titles are incorporated directly into the prompt, providing context that helps the model simultaneously identify and extract attribute-value pairs (See Figure~\ref{fig:generated_1}). In the two-step approach, the generated titles are first used to identify candidate attributes in the initial stage, refining the attribute identification process. (See Figure~\ref{fig:generated_2}) The identified attributes then guide the extraction of attribute-value pairs in the subsequent stage, aligning with the method defined in the two-step approach. This strategy simulates in-context learning by utilizing generated titles instead of retrieved labeled instances, effectively enhancing the model's capacity to handle the PAVI task without training data. The templates for generating pseudo inputs are illustrated in Figure~\ref{fig:self_generated_examples}.

\paragraph{Non-Parametric Knowledge.}
To enhance in-context learning, we leverage non-parametric knowledge by retrieving relevant examples from a labeled training set. Throughout our experiments, we distinguish between "titles" and "demonstrations": titles refer to product titles without any labels, while demonstrations refer to input/output pairs that include attribute-value annotations. Retrieving similar, unlabeled titles provides additional context that aids the model, especially in the two-step approach, by guiding attribute identification. We also retrieve labeled examples (demonstrations) from the training set to guide the model during inference. To select the most relevant demonstrations, we explore various retrieval strategies, beginning with baseline methods such as random selection, which provides an unbiased sampling, and lexical similarity using TF-IDF, which retrieves demonstrations based on the highest cosine similarity to the target input. Beyond these baselines, we employ more advanced retrieval techniques, including a pre-trained dense retriever based on the T5 model (T5-base\footnote{\url{https://huggingface.co/google-t5/T5-base}}) \cite{raffel2020exploring}, and a fine-tuned dense retriever specifically adapted for the PAVI task (T5-ft). We also experiment with varying the number of retrieved examples (1, 3, or 5) to assess their impact on performance.

The fine-tuned dense retriever is a sequence-to-sequence T5 model trained on the training set to generate attribute-value pairs given the product title. Given product title $x$, the model is trained to return a set of attribute-value pairs $y = \{\langle a_1, v_1 \rangle, \langle a_2, v_2 \rangle, \ldots, \langle a_k, v_k \rangle\}$, where $k$ is the number of attribute-value pairs associated with the product. During training, the model’s encoder-decoder architecture is used, but after training, we use only the fine-tuned encoder to produce dense embeddings of product titles. These embeddings capture task-specific patterns of attribute-value identification learned during training.. Cosine similarity is then used to retrieve the most relevant examples that align with the input data. The overall retrieval method is illustrated in Figure \ref{fig:method}.

\subsubsection{Instruction Fine-Tuning}

We fine-tune LLaMA-3 \cite{dubey2024LLaMA}, Mistral \cite{jiang2023mistral}, and OLMo \cite{groeneveld2024OLMo} models on the training sets of OA-Mine and AE-110k to enhance their performance on the PAVI task. The fine-tuning sets are created by formatting the records in the training datasets according to specific prompt templates, which generate role descriptions, task descriptions, and task inputs for each training record, as shown in Figure \ref{fig:one-step-zs}. The task output for each record contains the corresponding attribute-value pairs. Fine-tuning is conducted on all three models LLaMA, Mistral, and OLMo using Parameter-Efficient Fine-Tuning (PEFT) \cite{han2024parameter} with Low-Rank Adaptation (LoRA) \cite{hu2021lora}. Each model is fine-tuned for three epochs using all available training data. The specific configurations used for fine-tuning are detailed in Appendix \ref{app:hyper}.

\section{Experimental Settings} \label{sec: settings}

In this section, we outline the experimental settings used to evaluate the performance of our proposed methods for Product Attribute Value Identification.

\paragraph{Datasets.}

We use two real-world datasets:
\begin{itemize}[leftmargin=*,noitemsep,nolistsep]
\item{AE-110k}: The AE-110k dataset \cite{xu2019scaling} is a comprehensive collection of product data from the AliExpress Sports \& Entertainment category, containing 110,484 examples represented as triples: \{title, attribute, value\}. Each example includes a product title, an attribute, and its corresponding value, with NULL denoted for missing values. We removed instances with NULL values or entries containing invalid characters (e.g., '-' or '/'), resulting in a refined dataset of 109,957 valid triples. This dataset encompasses 39,505 products, 2,045 unique attributes, and 10,977 unique values, spread across 10 distinct product categories. Each category includes up to 400 product offers and features between 6 to 17 known attributes.

\item{OA-Mine}: The OA-Mine dataset \cite{zhang2022oa} is a human-annotated collection comprising 1,943 product entries across 10 product categories. Each category includes up to 200 product offers, featuring between 8 to 15 attributes, resulting in a total of 51 unique attributes. This dataset is used as-is without additional preprocessing.
\end{itemize}

\noindent Since there are no standard splits available for both AE-110k and OA-Mine, we randomly split both datasets into 80\% for training and 20\% for testing, ensuring that the split is stratified by product categories. The statistical details of both datasets are in Appendix \ref{app:datasets}.

\paragraph{Large Language Models.} This study evaluates three open-source LLMs: LLaMA-3\footnote{\url{https://huggingface.co/meta-LLaMA/Meta-LLaMA-3.1-8B-Instruct}} \cite{dubey2024LLaMA}, Mistral\footnote{\url{https://huggingface.co/mistralai/Mistral-7B-Instruct-v0.3}} \cite{jiang2023mistral}, and OLMo\footnote{\url{https://huggingface.co/allenai/OLMo-7B-Instruct-hf}} \cite{groeneveld2024OLMo}. Each model contains approximately 7 billion parameters, except for LLaMA-3, which has 8 billion parameters. These models are selected for their balance of performance and computational efficiency, making them suitable for practical deployment scenarios. An overview of the exact model names, sizes, and the number of GPUs used during evaluation are detailed in Appendix \ref{app:hyper}.

\paragraph{Evaluation Metrics.}
Following previous works \cite{yang2022mave,shinzato2023unified,roy2024exploring}, we use precision ($P$), recall ($R$), and $F_1$ score as evaluation metrics. The datasets may contain missing attribute-value pairs that the model might generate. To reduce the impact of such missing attribute-value pairs, following previous works \cite{shinzato2023unified,roy2024exploring}, we discard predicted attribute-value pairs if there are no ground truth labels for the generated attributes.
% Specifically, for each annotation, the model's predictions can fall into three categories: predicting the correct attribute but with an incorrect value (\textbf{WV}), failing to predict the attribute altogether, i.e., missing attribute (\textbf{MA}), or correctly predicting both the attribute and its value (\textbf{CV}). The precision, recall, and $F_1$ score are calculated as follows:

% \begin{small}
%  \begin{equation*}
% P = \frac{CV}{CV+WV} 
% \hspace{5mm}
% R = \frac{CV}{CV+MA}
% \hspace{5mm}
% F_1 = 2\times\frac{P \times R}{P + R}
% \end{equation*}
% \end{small}

\section{Results and Discussions}
\subsection{Zero-shot Results}

\begin{table}
\centering
\footnotesize
\resizebox{\columnwidth}{!}{
\begin{tabular}{|l| l l l l l|}
\hline
Dataset & Strategy & Model & $P$ & $R$ & $F_1$ \\
\hline
\multirow{12}{*}{AE-110k} 
 & \multirow{3}{*}{1-step} & LLaMA-3-8b & 41.56 & 2.42 & 4.58 \\
 &  & Mistral-7b & 54.21 & 10.22 & 17.20 \\
 &  & OLMo-7b & 34.22 & 6.41 & 10.79 \\
\cline{2-6} 
 & \multirow{3}{*}{1-step + SG} & LLaMA-3-7b & 44.98 & 2.51 & 4.75 \\
 &  & Mistral-7b & 57.96 & 11.00 & 18.48 \\
 &  & OLMo-7b & 43.25 & 10.68 & 17.13 \\
\cline{2-6} 
 & \multirow{3}{*}{2-step} & LLaMA-3-7b & 46.78 & 6.93 & 12.07 \\
 &  & Mistral-7b & \textbf{69.29}  & \textbf{18.32} & \textbf{28.97} \\
 &  & OLMo-7b & 51.46 & 7.60 & 13.24 \\
\cline{2-6} 
 & \multirow{3}{*}{2-step + SG} & LLaMA-3-8b & 42.52 & 10.93 & 17.37 \\
 &  & Mistral-7b & 67.65 & 16.48 & 26.51 \\
 &  & OLMo-7b & 43.05 & 7.54 & 12.83 \\
\hline
\multirow{12}{*}{OA-Mine} 
 & \multirow{3}{*}{1-step} & LLaMA-3-8b & 66.66 & 17.58 & 27.82 \\
 &  & Mistral-7b & \textbf{72.00} & 17.61 & 28.30 \\
 &  & OLMo-7b & 43.84 & 6.94 & 11.98 \\
\cline{2-6} 
 & \multirow{3}{*}{1-step + SG} & LLaMA-3-8b & 65.62 & 17.23 & 27.30  \\
 &  & Mistral-7b & 71.07 & 18.17 & 28.94 \\
 &  & OLMo-7b & 49.31 & 9.29 & 15.63 \\
\cline{2-6} 
 & \multirow{3}{*}{2-step} & LLaMA-3-8b & 65.37 & 20.86 & 31.64 \\
 &  & Mistral-7b & 67.21 & 19.96 & 30.78 \\
 &  & OLMo-7b & 21.40 & \textbf{23.80} & 22.54 \\
\cline{2-6} 
 & \multirow{3}{*}{2-step + SG} & LLaMA-3-8b & 63.32 & 22.70 & 33.42 \\
 &  & Mistral-7b & 69.13 & 18.96 & \textbf{29.75}  \\
 &  & OLMo-7b & 23.42 & 25.12 & 24.24 \\
\hline
\end{tabular}
}
\caption{Zero-shot results across AE-110k and OA-Mine datasets. 1-step refers to the one-step approach, and 2-step refers to the two-step approach. SG stands for self-generated examples. The best results for each dataset are highlighted in bold.}
\label{tb:res-zs}
\end{table}

Table \ref{tb:res-zs} shows the performance of the evaluated models in zero-shot settings across the AE-110k and OA-Mine datasets. The results indicate that the two-step approach consistently outperforms the one-step approach in both datasets due to its sequential prompting strategy. This approach guides the model through a sequential process, first identifying attributes and then extracting corresponding values. This allows the model to focus on each sub-task more effectively, thereby improving overall accuracy. For example, on the AE-110k dataset, Mistral achieves an $F_1$ score of 28.97 with the two-step approach compared to 17.20 with the one-step approach, demonstrating a significant performance gain.
Incorporating self-generated examples, through parametric knowledge, generally enhances performance, particularly in the one-step approach, although the impact varies across models and datasets. For instance, in the AE-110k dataset, OLMo shows a significant improvement in the one-step approach when self-generated examples are used, with its $F_1$ score increasing from 10.79 to 17.13. This improvement underscores the benefit of self-generated data in guiding the models predictions, though gains are not consistent; in some cases, improvements are slight or inconsistent, indicating that the effectiveness of self-generated examples may depend on the specific characteristics of the model and dataset. 
When comparing the models, Mistral consistently achieves higher precision, recall, and $F_1$ scores compared to both LLaMA and OLMo across all strategies. This highlights Mistrals overall robustness in zero-shot settings, particularly with the two-step method. These findings reinforce the advantage of the two-step approach as an effective extraction process in zero-shot scenarios, particularly when enhanced with self-generated data using parametric knowledge.

% \begin{itemize}
%     \item Two-step approach outperforms one-step approach in zero-shot settings.
%     \item Self-generation helps and two-step still outperforms one-step.
% \end{itemize}
% \emph{Should add the prompt template for self-generation here (possibly in Appendix)}

\subsection{In-context Learning with Titles}

\begin{table}
\centering
\resizebox{\columnwidth}{!}{
\begin{tabular}{|l|lllll|}
\hline
Dataset & Strategy & Model & P & R & $F_1$ \\ \hline
\multirow{6}{*}{AE-110k} & \multirow{3}{*}{2-step} & LLaMA-3-8b & 46.78 & 6.93 & 12.07 \\
 &  & Mistral-7b & 69.29 & \textbf{18.32} & \textbf{28.97} \\
 &  & OLMo-7b & 51.46 & 7.60 & 13.24 \\ \cline{2-6} 
 & \multirow{3}{*}{2-step + Titles} & LLaMA-3-8b & 43.29 & 9.81 & 16.00 \\
 &  & Mistral-7b & \textbf{69.35} & 17.41 & 27.84 \\
 &  & OLMo-7b & 42.91 & 7.55 &  12.84\\ \hline
\multirow{6}{*}{OA-Mine} & \multirow{3}{*}{2-step} & LLaMA-3-8b & 65.37 & 20.86 & 31.64 \\
 &  & Mistral-7b & \textbf{67.21} & 19.96 & 30.78 \\
 &  & OLMo-7b & 21.40 & \textbf{23.80} & 22.54 \\ \cline{2-6}
 & \multirow{3}{*}{2-step + Titles} & LLaMA-3-8b & 63.58 & 22.87 & \textbf{33.64} \\
 &  & Mistral-7b & 66.98 & 18.70 & 29.23 \\
 &  & OLMo-7b & 20.05 & 23.76 & 21.75 \\ \hline
\end{tabular}
}
\caption{Performance of LLaMA, Mistral, and OLMo in the two-step approach using only retrieved titles.}

\label{tb:res-titles}
\end{table}

% \begin{itemize}
%     \item Helps for LLLaMA but not so much for mistral (We do not know why for now)
%     \item Perfromance is comparable to self-generation)
%     \item OLMo should also provide some insights compared to the two models.
%     \item \emph{We only use a lexical retriever here, we should probably add random and dense retrievals}
% \end{itemize}
% \emph{Should add the prompt template of how we do this in Appendix}

Table \ref{tb:res-titles} presents the results of using only product titles retrieved with TF-IDF in the two-step approach across the AE-110k and OA-Mine datasets\footnote{We only show results with TF-IDF as other retrieval strategies yielded similar performance, indicating that the retrieval method had a minimal impact when using only product titles.}. The results show that incorporating product titles as the primary context generally improves performance for the LLaMA model, particularly on the OA-Mine dataset, where the $F_1$ score increases from 31.64 to 33.64. This suggests that LLaMA benefits more from retrieving titles as additional context. However, this strategy does not significantly benefit the Mistral and OLMo models, which experience slight decreases in performance on both datasets when titles are used. When comparing these results to those obtained using self-generated examples, it becomes clear that using only titles, whether they are generated by the model or retrieved from the training set, can effectively improve performance. This approach is especially important in scenarios where labeled data is not available, and only title texts are provided. This method can be highly beneficial in practical applications like e-commerce, where product titles are available, but annotated examples may not always be present.

\subsection{In-context Learning with Demonstrations}

\begin{table}
\scriptsize
\centering
\resizebox{\columnwidth}{!}{
\begin{tabular}{|l|lllll|}
\hline
Dataset & Strategy & Model & P & R & $F_1$ \\ \hline
\multirow{12}{*}{AE-110k} & \multirow{3}{*}{1-step} & LLaMA-3-8b & 41.56 & 2.42 & 4.58 \\ 
 & & Mistral-7b & 54.21 & 10.22 & 17.20 \\ 
 & & OLMo-7b & 34.22 & 6.41 & 10.79 \\ \cline{2-6} 
 & \multirow{3}{*}{1-step + Rnd} & LLaMA-3-8b & 66.2 & 19.45 & 30.17 \\
 &  & Mistral-7b & 72.17 & 20.03 & 31.35 \\ 
 &  & OLMo-7b & 60.48 & 15.27 & 24.38 \\ \cline{2-6} 
 & \multirow{3}{*}{1-step + TF-IDF} & LLaMA-3-8b & 81.58 & 60.58 & 69.64 \\
 &  & Mistral-7b & 81.06 & 55.46 & 65.86 \\ 
 &  & OLMo-7b & 81.75 & 25.31 & 38.65 \\ \cline{2-6}
 & \multirow{3}{*}{1-step + T5-base} & LLaMA-3-8b & 82.15 & 62.01 & 70.67 \\
 &  & Mistral-7b & \textbf{84.98} & 52.37 & 64.80 \\ 
 &  & OLMo-7b & 82.01 & 38.86 &  52.73\\ \cline{2-6}
 & \multirow{3}{*}{1-step + T5-ft} & LLaMA-3-8b & 83.24 & \textbf{65.53} & \textbf{73.33} \\
 &  & Mistral-7b & 84.70 & 55.18 & 66.80 \\ 
 &  & OLMo-7b &81.91 & 42.57  &  56.02\\ \hline
\multirow{12}{*}{OA-Mine} & \multirow{3}{*}{1-step} & LLaMA-3-8b & 66.66 & 17.58 & 27.82 \\
 &  & Mistral-7b & 72.00 & 17.61 & 28.30 \\
 &  & OLMo-7b & 43.84 & 6.94 & 11.98 \\ \cline{2-6} 
 & \multirow{3}{*}{1-step + Rnd} & LLaMA-3-8b & 67.62 & 26.41 & 37.99 \\
 &  & Mistral-7b & 68.17 & 25.58 & 37.20 \\ 
 &  & OLMo-7b & 71.17 & 18.27 & 29.08 \\ \cline{2-6} 
 & \multirow{3}{*}{1-step + TF-IDF} & LLaMA-3-8b & 73.72 & 70.05 & 71.84 \\
 &  & Mistral-7b & 75.06 & 65.49 & 69.95 \\ 
 &  & OLMo-7b & 73.70 & 35.14 & 48.94 \\ \cline{2-6}
 & \multirow{3}{*}{1-step + T5-base} & LLaMA-3-8b & \textbf75.95 & 71.05 & 73.42 \\
 &  & Mistral-7b & \textbf{82.15} & 62.01 & 70.60 \\ 
 &  & OLMo-7b & 78.08 & 41.27 & 54.00 \\ \cline{2-6}
 & \multirow{3}{*}{1-step + T5-ft} & LLaMA-3-8b & 75.13 & \textbf{71.91} & \textbf{73.48} \\
 &  & Mistral-7b & 75.89 & 68.14 & 71.80 \\ 
 &  & OLMo-7b & 75.63 & 48.64 & 59.20 \\ \hline
\end{tabular}
}
\caption{Performance comparison of LLaMA, Mistral, and OLMo using various in-context demonstration strategies. Rnd stands for random.}

\label{tb:in-context-results}
\end{table}

% \begin{itemize}
%     \item Titles and labels provide better results as compared to just titles.
%     \item Random examples already improves over just titles.
%     \item Lexical examples improve over random examples.
%     \item Semantical retriever get marginal improvements over lexical retriever.
%     \item Ft retriever improves over base semantic retriever.
% \end{itemize}

\paragraph{Demonstration Selector.} Table \ref{tb:in-context-results} presents the results of incorporating in-context demonstrations using various retrieval strategies in the one-step approach across the AE-110k and OA-Mine datasets. The results indicate that using titles along with attribute-value labels significantly outperforms using only product titles. Randomly selected examples already show a notable improvement over the baseline one-step approach, indicating that even non-specific in-context demonstrations provide valuable contextual information that enhances model performance. For instance, Mistral's $F_1$ score on the AE-110k dataset increases from 17.20 in the one-step approach to 31.35 when random examples are used. Lexical retriever using TF-IDF, further improves performance over random examples, by retrieving examples that are more relevant to the target context based on lexical similarity. For example, Mistral achieves an $F_1$ score of 65.86 on the AE-110k dataset with TF-IDF, a significant improvement over randomly retrieved examples. Using T5-base as a dense retriever shows marginal gains over TF-IDF, as the T5-base model captures semantic contextual information. For instance, OLMo improves from an $F_1$ score of 38.65 with TF-IDF to 52.73 with T5-base on the AE-110k dataset, highlighting the benefits of using semantical representations. Finally, our proposed fine-tuned T5 retriever (T5-ft) demonstrates the highest performance gains across both datasets, outperforming the base T5 model. The fine-tuning process aligns the retriever with the specific attribute-value identification task, leading to more precise and contextually relevant example selection. For example, on the OA-Mine dataset, LLaMA $F_1$ score reaches 73.33 with the fine-tuned T5 retriever, compared to 70.67 with the T5-base retriever.

\begin{table}
\centering
\resizebox{\columnwidth}{!}{
\begin{tabular}{|l|lllll|}
\hline
Dataset & Strategy & Model & 1 Example & 3 Examples & 5 Examples \\ \hline

\multirow{6}{*}{AE-110k} & \multirow{3}{*}{1-step + T5-base} & LLaMA-3-8b & 70.67 & 74.34 & 75.62 \\
 & & Mistral-7b & 64.80 & 56.33 & 50.54 \\
 & & OLMo-7b & 52.73 & 63.69& 69.85 \\ \cline{2-6} 

 & \multirow{3}{*}{1-step + T5-ft} & LLaMA-3-8b & 73.33 & 75.25 & \textbf{77.36} \\
     &  & Mistral-7b & 66.80 & 58.44 & 52.60 \\
 &  & OLMo-7b & 56.02 & 70.31 & 71.86 \\ \hline

\multirow{6}{*}{OA-Mine} & \multirow{3}{*}{1-step + T5-base} & LLaMA-3-8b & 73.42 & 74.10 & 74.74 \\ 
 & & Mistral-7b & 70.60 & 64.44 & 65.60 \\
 & & OLMo-7b & 54.00 & 58.21 & 60.13 \\ \cline{2-6} 
 
 & \multirow{3}{*}{1-step + T5-ft} & LLaMA-3-8b & 73.48 & 74.02 & \textbf{75.24} \\
 &  & Mistral-7b & 71.80 & 63.70 & 64.55 \\
 &  & OLMo-7b & 59.20 & 61.35 & 66.10  \\ \hline
 
\end{tabular}
}
\caption{Impact of varying the number of in-context demonstrations on the performance ($F_1$) of the models for dense retrievers.}
\label{tb:res-number-demos}
\end{table}

% \begin{figure}
%     \centering \includegraphics[width=1\columnwidth]{figures/results.eps}
%     \caption{}
%     \label{fig:number-demons}
% \end{figure}

% \begin{itemize}
%     \item Increasing the number of examples improves the performance across datasets and models.
% \end{itemize}

\paragraph{Number of Demonstrations.} Table \ref{tb:res-number-demos} shows the effect of varying the number of in-context demonstrations (1, 3, and 5) on the performance of LLaMA, Mistral, and OLMo using the T5-base and T5-ft retrievers. The results indicate that increasing the number of demonstrations generally improves performance across all models and datasets, although the extent of improvement varies by model and retrieval strategy. For LLaMA, performance consistently improves as the number of retrieved examples increases, with the highest $F_1$ scores observed when five examples are used. On the AE-110k dataset, LLaMA achieves an $F_1$ score of 77.36 with T5-ft and five examples, highlighting the benefits of additional contextual information. OLMo also shows positive trends, particularly with T5-ft, achieving its best performance at five examples. However, for Mistral, the performance gains are less consistent, and in some cases, performance decreases as more examples are added. This suggests that Mistral may be more sensitive to the quality and relevance of the retrieved examples, particularly when using the T5-base retriever. Overall, these findings suggest that LLaMA shows the most consistent gains, underscoring its robustness when utilizing multiple in-context demonstrations, particularly with the fine-tuned retriever.

\begin{table}
\centering
\resizebox{\columnwidth}{!}{
\begin{tabular}{|l|l|lll|}
\hline
Dataset & Model & 1 Example & 3 Examples & 5 Examples \\ \hline

\multirow{3}{*}{AE-110k} 
 & LLaMA-3-8b & 72.11 & 74.21 & \textbf{75.51} \\
 & Mistral-7b & 76.45 & 57.69 & 49.47 \\
 & OLMo-7b & 55.44 & 68.16 & 71.59 \\ \hline
 
\multirow{3}{*}{OA-Mine} 
 & LLaMA-3-8b & 73.76 & 74.68 & \textbf{75.00} \\
 & Mistral-7b & 70.82 & 65.38 & 66.10 \\
 & OLMo-7b & 59.73 & 60.86 & 61.71 \\ \hline
 
\end{tabular}
}
\caption{Domain transfer results showing $F_1$ scores for LLaMA, Mistral, and OLMo when using T5 retrievers fine-tuned on one dataset and tested on another.}
\label{tb:domain-res}
\end{table}

\paragraph{Domain Transfer.}
Table \ref{tb:domain-res} shows the results of domain transfer experiments, where the T5 retriever is fine-tuned on one dataset and tested on another. The results highlight that domain transfer can have varied effects depending on the model and dataset. Notably, transferring the T5 retriever fine-tuned on AE-110k to OA-Mine generally results in comparable or improved performance relative to in-domain fine-tuning, especially for the LLaMA model. For example, LLaMA achieves an $F_1$ score of 75.00 on OA-Mine with the cross-domain T5 retriever, nearly matching its in-domain performance of 75.24. Interestingly, Mistral shows improved performance when transferring from OA-Mine to AE-110K, achieving an $F_1$ score of 76.45 on AE-110k with the cross-domain retriever compared to 66.80 using a single in-context demonstration in the in-domain setting. However, consistent with in-domain results, Mistral's performance decreases with more retrieved examples. OLMo demonstrates mixed results in the cross-domain experiment, with some performance improvements, particularly on AE-110k, where it reaches an $F_1$ score of 71.59 when retrieving five examples. Overall, these findings underscore that retriever domain transfer is a valuable strategy, particularly in e-commerce where new product categories and domains frequently emerge.

% \begin{itemize}
%     \item Transfer from AE-110k to OA-mine performs better/comparable than in-domain OA-Mine
    
% \end{itemize}

\subsection{Instruction Fine-tuning}
\begin{table}
\centering
\resizebox{\columnwidth}{!}{
\begin{tabular}{|l|llll|}
\hline
Dataset & Model & $P$ & $R$ & $F_1$ \\ \hline

\multirow{3}{*}{AE-110k} 
& LLaMA-3-8b & \textbf{94.74} & \textbf{70.87} & \textbf{81.09} \\
& Mistral-7b & 92.89 & 66.72 & 77.66 \\
& OLMo-7b & 92.52 & 64.53 & 76.03 \\ \hline

\multirow{3}{*}{OA-Mine} 
& LLaMA-3-8b & 80.22 & \textbf{87.51} & \textbf{83.70} \\
& Mistral-7b & 76.54 & 69.08 & 72.62 \\
& OLMo-7b & \textbf{92.98} & 62.03 & 74.41 \\ \hline

\end{tabular}
}
\caption{Performance of LLaMA, Mistral, and OLMo on the AE-110k and OA-Mine datasets after instruction fine-tuning.}
\label{tb:res-ft}
\end{table}

Table \ref{tb:res-ft} shows that instruction fine-tuning significantly enhances model performance compared to in-context learning. LLaMA achieves the highest $F_1$ score on AE-110K (81.09), surpassing its previous best of 77.36 from in-context learning. This shows the impact of explicit task-specific fine-tuning. Mistral and OLMo also show improved $F_1$ scores of 77.66 and 76.03, respectively, indicating the effectiveness of fine-tuning across models. On OA-Mine, LLaMA achieves an $F_1$ score of 83.70, significantly outperforming its in-context results, while Mistral and OLMo also benefit from fine-tuning. These results suggest that while in-context learning can enhance performance by leveraging parametric and non-parametric knowledge, instruction fine-tuning offers a more direct adaptation to the task, leading to consistently higher performance.

\section{Conclusions} 
In this paper, we explored the capabilities of large language models (LLMs) for PAVI. We propose various approaches, including one-step and two-step prompt-based strategies. We proposed a dense demonstration
retriever based on a pre-trained T5 model. Our experiments show that the two-step approach consistently outperformed the one-step method in zero-shot settings. With available training data, using our fine-tuned retriever boosts results for in-context learning, and instruction fine-tuning provides the highest performance gains.

\section*{Limitations}
While our study demonstrates the effectiveness of LLMs for Product Attribute Value Identification (PAVI), it also has several limitations. First, the datasets used in our experiments, AE-110K and OA-Mine, lack standard splits, so we randomly split the data as detailed in Section \ref{sec: settings}. Second, our evaluation metrics do not penalize over-generated attribute-value pairs, potentially omitting valid extractions due to incomplete annotations. As future work, we plan to develop methods to assess these newly generated attributes automatically. Third, our evaluation focused solely on open-source LLMs such as LLaMA, Mistral, and OLMo, and did not include proprietary models, such as GPT-4, which may perform differently. In the future, we plan to evaluate proprietary LLMs alongside open-source models to explore their potential advantages in performance, generalization, and efficiency. 

\bibliography{custom}

\appendix

\section{Prompt Templates}

\begin{figure}
    \centering
\includegraphics[width=1\linewidth]{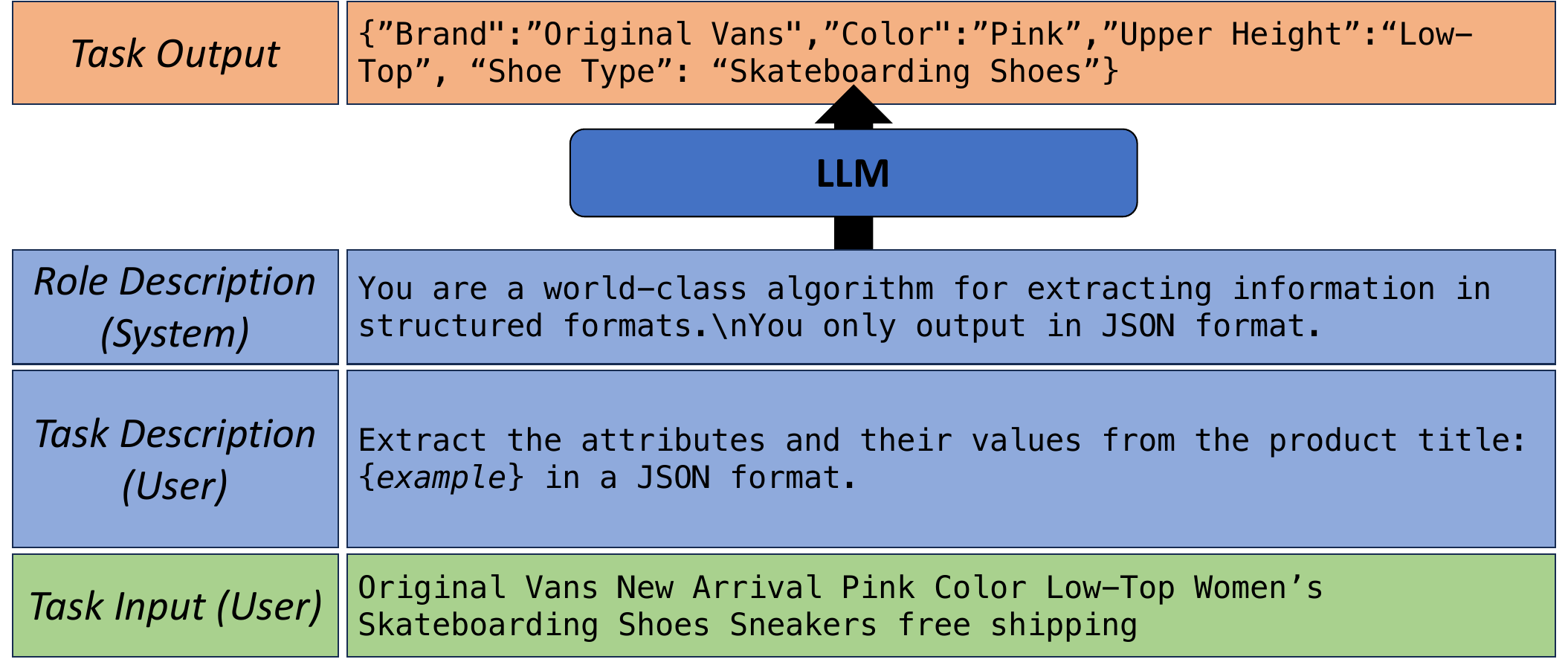}
\caption{Zero-shot prompt template for the one-step approach.}
    \label{fig:one-step-zs}
\end{figure}

\begin{figure*}
    \centering
    \begin{subfigure}[b]{0.49\textwidth}
        \centering
        \includegraphics[width=\textwidth]{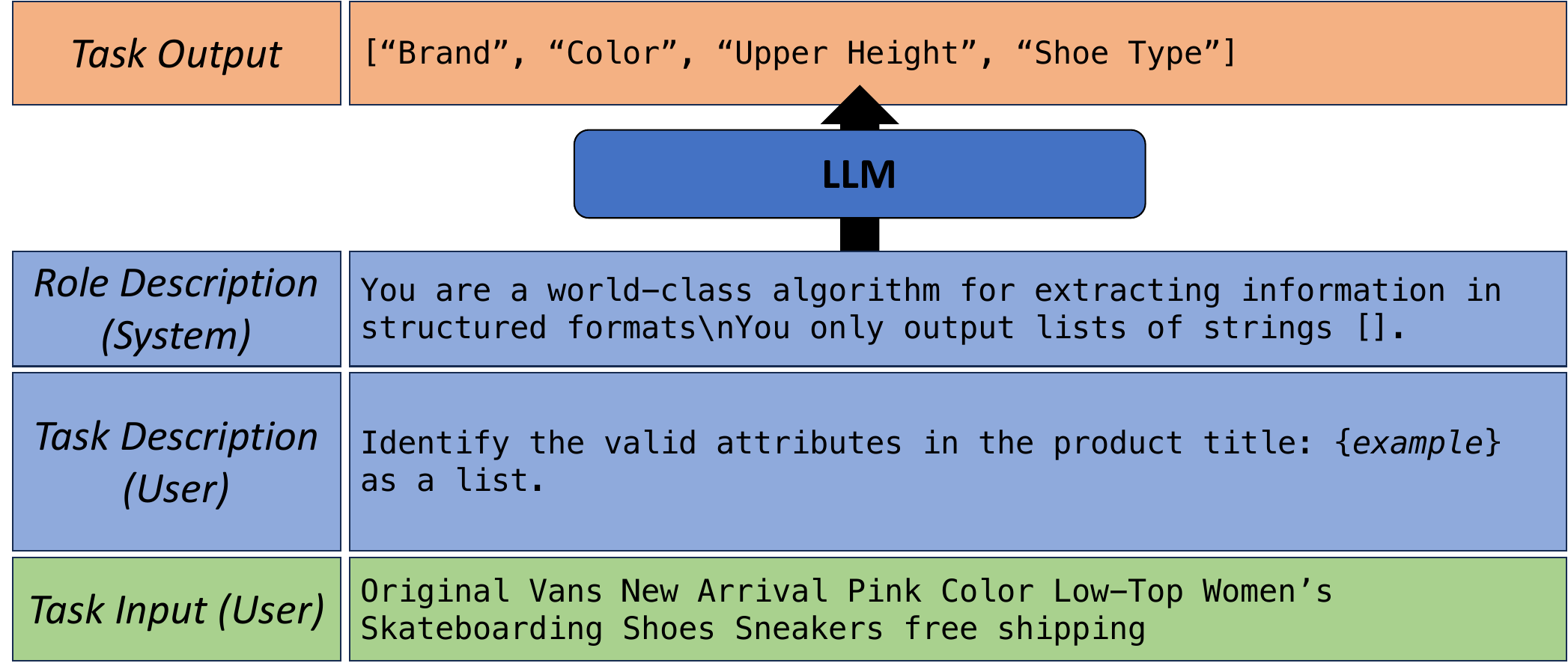}
        \caption{In the first stage, the model is prompted to identify the attributes.}
        \label{fig:figure1}
    \end{subfigure}
    \hfill
    \begin{subfigure}[b]{0.49\textwidth}
        \centering
        \includegraphics[width=\textwidth]{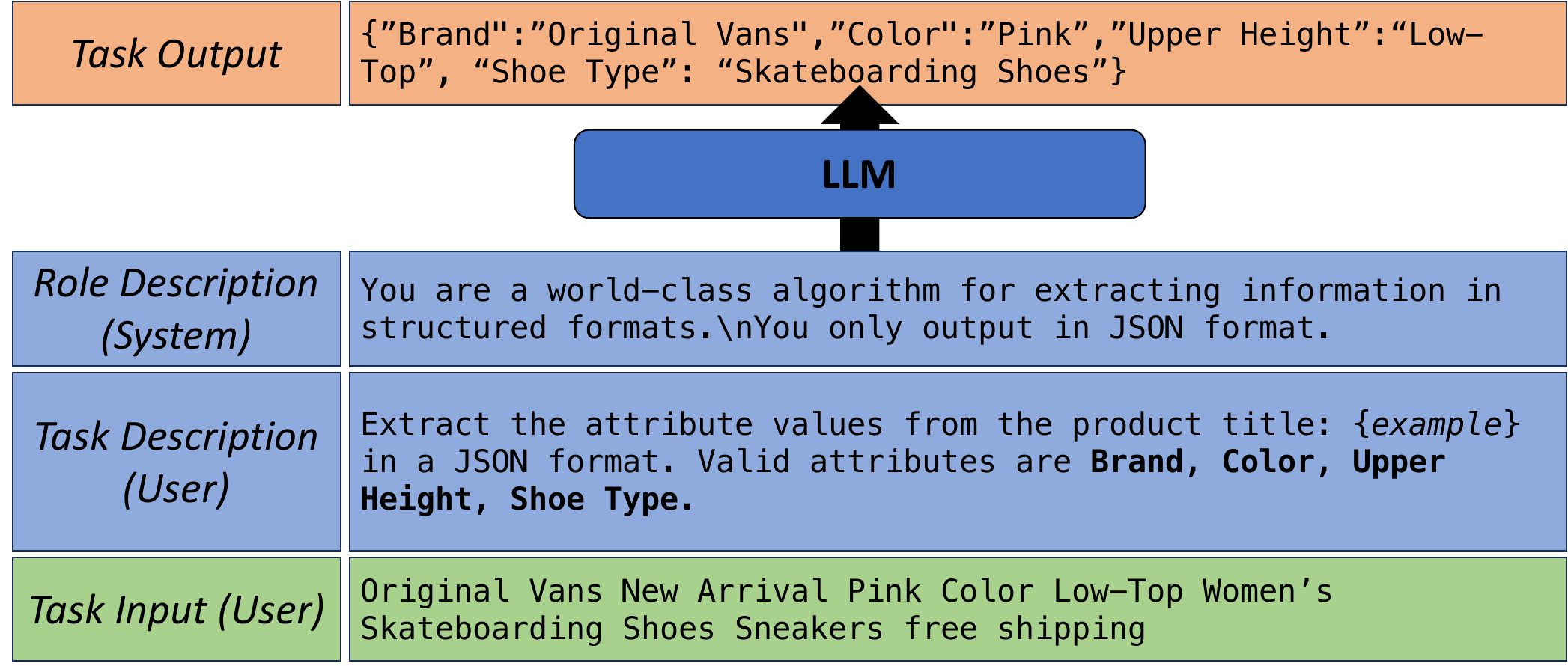}
        \caption{In the second stage, the identified attributes are used to prompt the model to extract the corresponding values.}
        \label{fig:figure2}
    \end{subfigure}
    \caption{Zero-shot prompt template for the two-step approach.}
    \label{fig:two-step-zs}
\end{figure*}

\begin{figure}
    \centering
\includegraphics[width=1\linewidth]{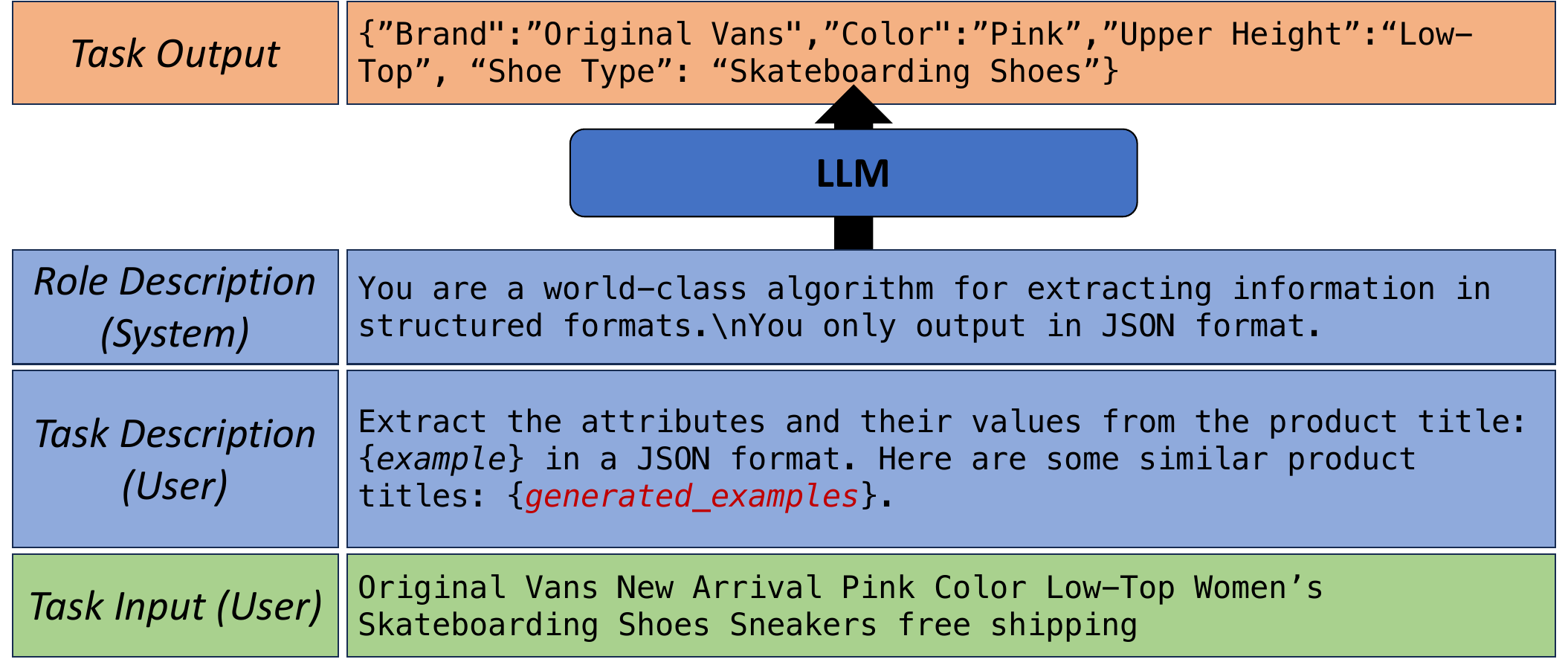}
\caption{Template illustrating how self-generated product titles are used in the one-step approach.}
    \label{fig:generated_1}
\end{figure}

\begin{figure}
    \centering
\includegraphics[width=1\linewidth]{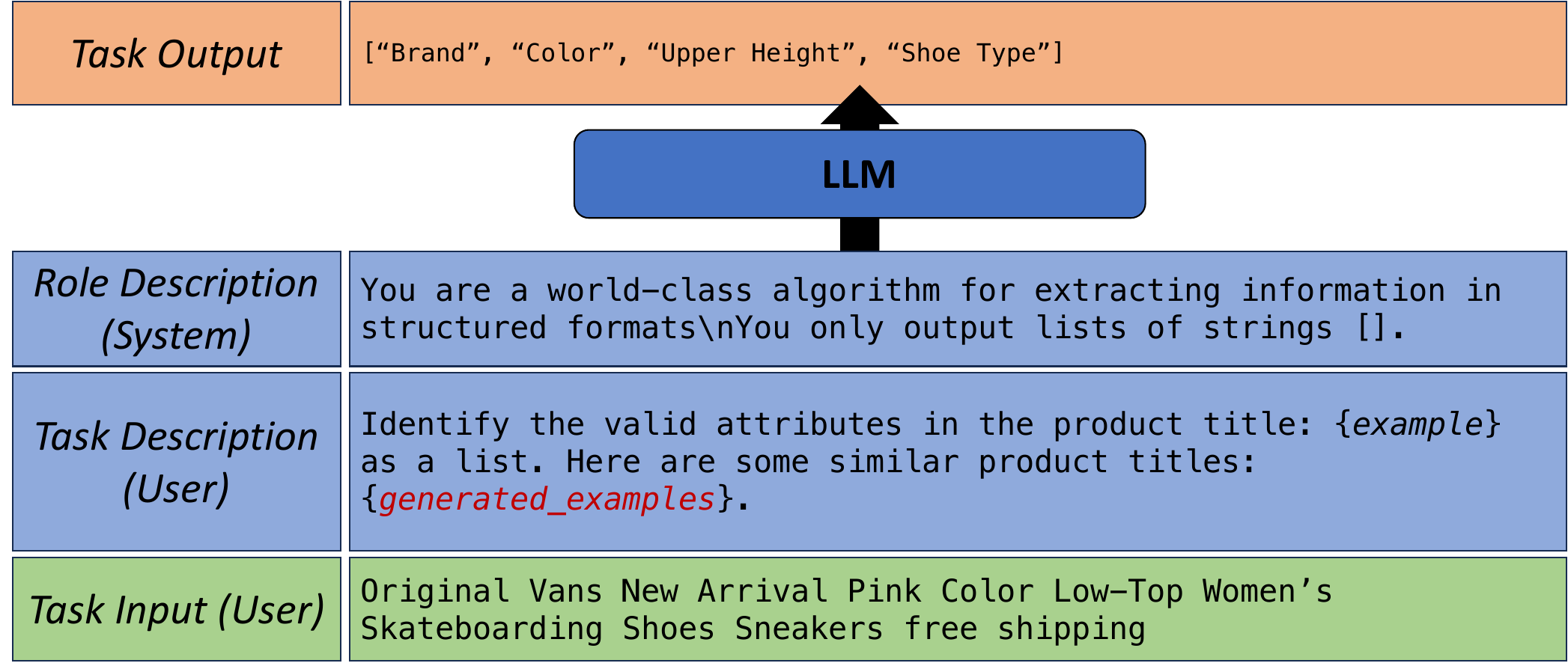}
\caption{Template illustrating how self-generated product titles are used in the first step (attribute identification) of the two-step approach.}
    \label{fig:generated_2}
\end{figure}

\begin{figure}
    \centering
\includegraphics[width=1\linewidth]{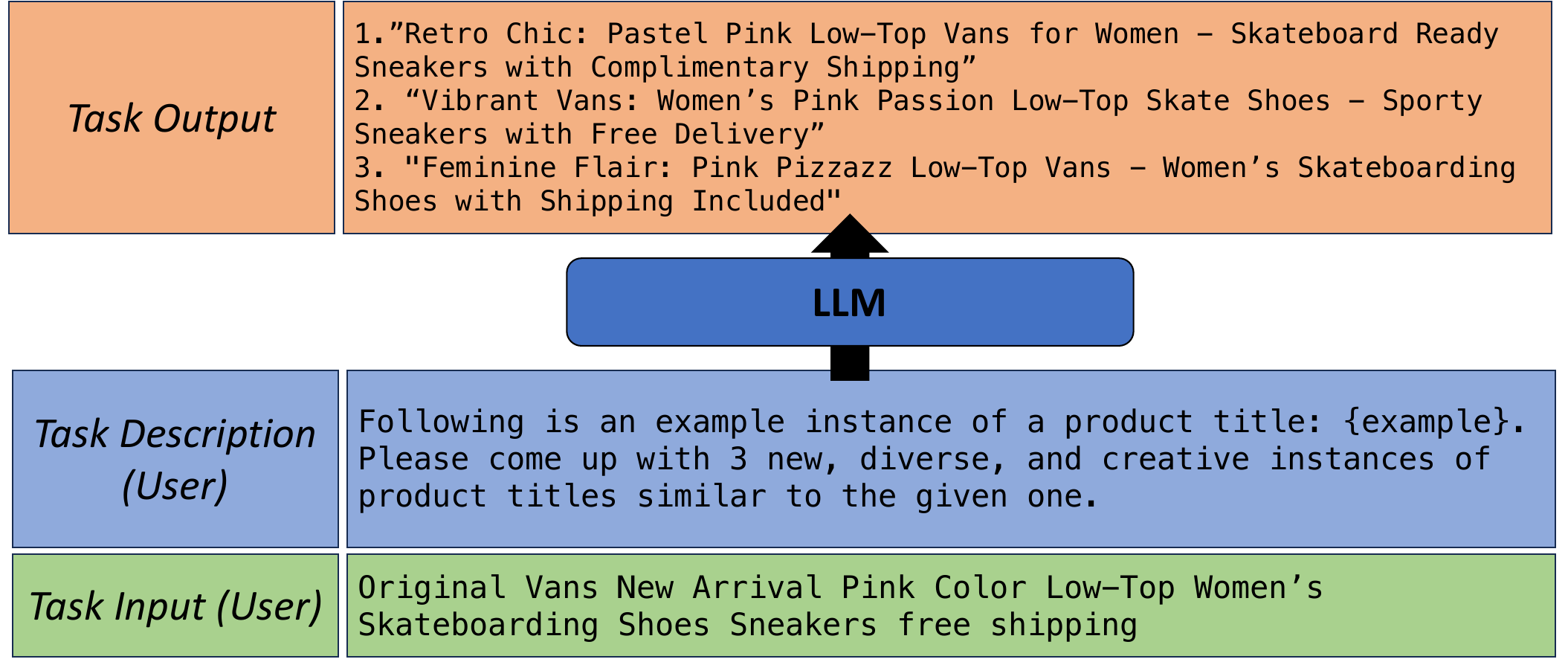}
\caption{Template illustrating the self-generated examples approach. Given an example product title, the model generates diverse pseudo titles, which are then used to either directly extract attribute-value pairs (one-step) or first identify attributes and then extract values (two-step).}
    \label{fig:self_generated_examples}
\end{figure}

\section{Datasets} \label{app:datasets}

\begin{table}
\centering
\resizebox{\columnwidth}{!}{
\begin{tabular}{|ccc|}

\hline
\textbf{Counts} & AE-110k & OA-Mine \\ \hline

\# products & 39,505 & 1,943 \\
\# attribute-value pairs & 88,915 & 11,008  \\
\# unique categories & 10 & 10  \\
\# unique attributes & 2,045 & 51  \\
\# unique values & 10,977 & 5,201  \\ \hline
\end{tabular}
}
\caption{Statistics of AE-110k, OA-Mine, and MAVE datasets.}
\label{tb:statistics}
\end{table}

Table \ref{tb:statistics} provides statistical details of the AE-110K \cite{xu2019scaling} and OA-Mine \cite{zhang2022oa} datasets used in our experiments, including the number of products, attribute-value pairs, unique categories, attributes, and values.

\section{Implementation Details} \label{app:hyper}

\begin{table}
\centering
\resizebox{\columnwidth}{!}{%
\begin{tabular}{|l|l|}
\hline
\multicolumn{2}{|c|}{\textbf{LoRA Configuration}} \\ 
\hline
LoRA Alpha                              & 128                       \\ 
LoRA Dropout                            & 0.05                      \\ 
Rank ($r$)                              & 256                       \\ 
Bias                                    & None                      \\ 
Target Modules                          & All Linear                \\ 
\hline
\multicolumn{2}{|c|}{\textbf{Training Configuration}} \\ 
\hline
Number of Training Epochs               & 3                         \\ 
Per Device Train Batch Size             & 1                         \\ 
Gradient Accumulation Steps             & 8                         \\ 
Learning Rate                           & 2e-4                      \\ 
Max Gradient Norm                       & 0.3                       \\ 
Warmup Ratio                            & 0.03                      \\ 
Max Sequence Length                     & 2048                      \\ 
\hline
\end{tabular}%
}
\caption{LoRA and Training Configuration for Instruction Fine-Tuning.}
\label{tb:lora_training_config}
\end{table}

\paragraph{Instruction Fine-tuning Hyper-parameters.} Table \ref{tb:lora_training_config} summarizes the LoRA and training configurations used for instruction fine-tuning in our experiments. The configurations include key parameters such as LoRA Alpha, dropout rate, training epochs, batch size, and learning rate. The hyper-parameters were chosen based on preliminary experiments and best practices from relevant literature. The table provides all necessary details for reproducing the experiments.

\begin{table}
\centering
\resizebox{\columnwidth}{!}{
\begin{tabular}{|llll|}
\hline
LLM &  Exact Name & Model Size & GPUs \\
\midrule
LLaMA-3 & Meta-LLaMA-3.1-8b-Instruct & 8B & 1 \\
Mistral & Mistral-7B-Instruct-v0.3 & 7B & 1 \\
OLMo & OLMo-7B-Instruct-hf & 7B & 1 \\
\hline
\end{tabular}
}
\caption{List of evaluated LLMs, including their exact names, sizes, and GPU usage during evaluation.}
\label{tb:models}
\end{table}

\paragraph{Large Language Models.} Table \ref{tb:models} provides details of the evaluated large language models (LLMs), including their exact names, sizes, and GPU usage during evaluation. All models, including LLaMA-3 (8B), Mistral (7B), and OLMo (7B), were run locally on a single GPU with a fixed temperature parameter of 0 to ensure consistency and minimize output variability. The experiments were conducted on a shared server with substantial computational resources, including 96 CPU cores, 512 GB RAM, and 8 NVIDIA A100 GPUs, although only one GPU was used per model to maintain a controlled evaluation environment.

\end{document}